%% file: main.tex
\documentclass{article}
\usepackage[T1]{fontenc}
\usepackage[utf8]{inputenc}
\usepackage{hyperref}
\usepackage{spconf,amsmath,graphicx}
\usepackage{float}
\usepackage{amsmath}
\usepackage{mathtools}
\usepackage[font=normalsize,skip=12pt]{caption}
\usepackage{flushend}
\usepackage[backend=biber,style=ieee,natbib=false,mincrossrefs=1000]{biblatex}
\usepackage{fancyhdr}
\AtBeginDocument{\toggletrue{blx@useprefix}}
\AtBeginBibliography{\togglefalse{blx@useprefix}}

\DeclareMathOperator*{\argmin}{arg\,min}

\newcommand{\KLD}[2]{D_{\mathrm{KL}}\left( #1 \left|\right| \, #2 \right)}

\title{Image-dependent local entropy models\\
for learned image compression}

\name{David Minnen, George Toderici, Saurabh Singh, Sung Jin Hwang, Michele Covell}
 
\address{Google, 1600 Amphitheatre Pkwy., Mountain View, CA 94043, USA}

\begin{document}

\maketitle

\begin{abstract}
\input{abstract.tex}
\end{abstract}


\fancypagestyle{firststyle}
{
   \fancyhf{}
   \lfoot{\textcopyright 2018 IEEE}
   \setcounter{page}{1}
   \cfoot{\thepage}
   \rfoot{ICIP 2018}
}

\thispagestyle{firststyle}

\fancyhf{}
\renewcommand{\headrulewidth}{0pt}
\setcounter{page}{1}
\cfoot{\thepage}

\input{intro.tex}
\input{method.tex}
\input{eval.tex}
\input{conclusion.tex}

\newpage

\printbibliography

\end{document}

%% file: abstract.tex
The leading approach for image compression with artificial neural networks (ANNs) is to learn a nonlinear transform and a fixed entropy model that are optimized for rate-distortion performance. We show that this approach can be significantly improved by incorporating spatially local, image-dependent entropy models. The key insight is that existing ANN-based methods learn an entropy model that is shared between the encoder and decoder, but they do not transmit any side information that would allow the model to adapt to the structure of a specific image. We present a method for augmenting ANN-based image coders with image-dependent side information that leads to a 17.8\% rate reduction over a state-of-the-art ANN-based baseline model on a standard evaluation set, and 70--98\% reductions on images with low visual complexity that are poorly captured by a fixed, global entropy model.

%% file: intro.tex
\section{Introduction}
\label{sec:intro}

Lossy image compression based on nonlinear transform coding with artificial neural networks (ANNs) is a research topic that has recently garnered significant interest~\cite{ToOMHwViMi16, AgMeTsCaTi17, BaLaSi17, RiBo17, ThShCuHu17, ToViJoHwMi17, minnen2017icip, JoViMiCoSi17}. One of the primary challenges faced by this approach is how to construct effective probability models used to entropy code the quantized symbols generated by the networks. Unlike conventional (non-learned) methods for lossy image compression, the output from deep neural networks does not have a known structure. So while a codec designer could make use of the ordered frequency responses of the DCT transform when constructing a quantization table, the codes generated by an optimized neural network may not have an obvious structure and will likely change each time the model is re-trained or the network architecture is modified.

\begin{figure}[t]
  \centering
  \includegraphics[width=\columnwidth]{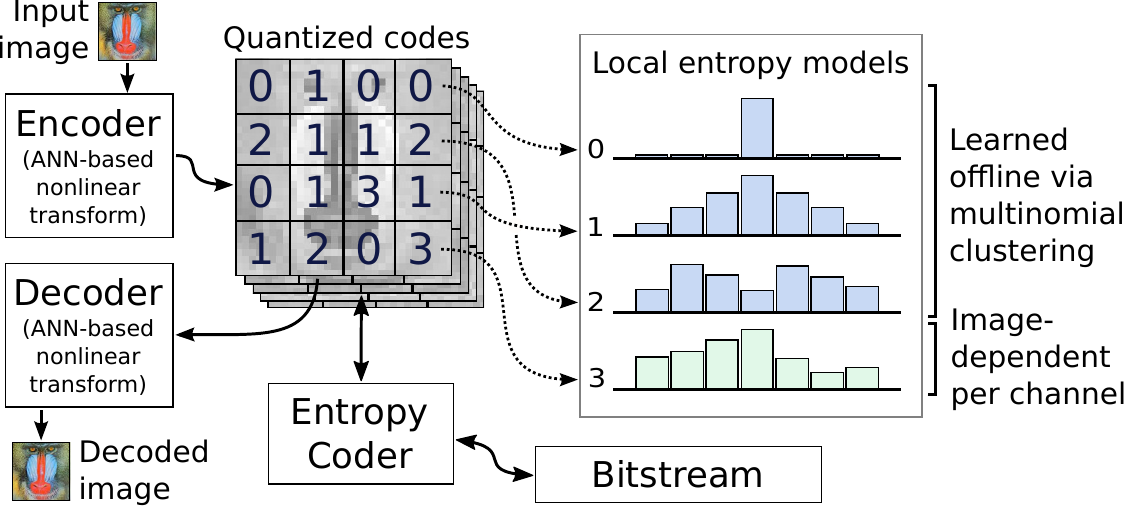}
  \caption{Our method uses a combination of pre-learned and image-dependent multinomials to improve the (lossless) compression rate of the quantized codes generated by an ANN-based nonlinear transform. The encoder selects the distribution with the lowest cross-entropy for each code tile, which provides local adaptation and higher compression rates.}
  \label{fig:method}
\end{figure}

\begin{figure*}[t]
  \centering
  \includegraphics[width=\textwidth]{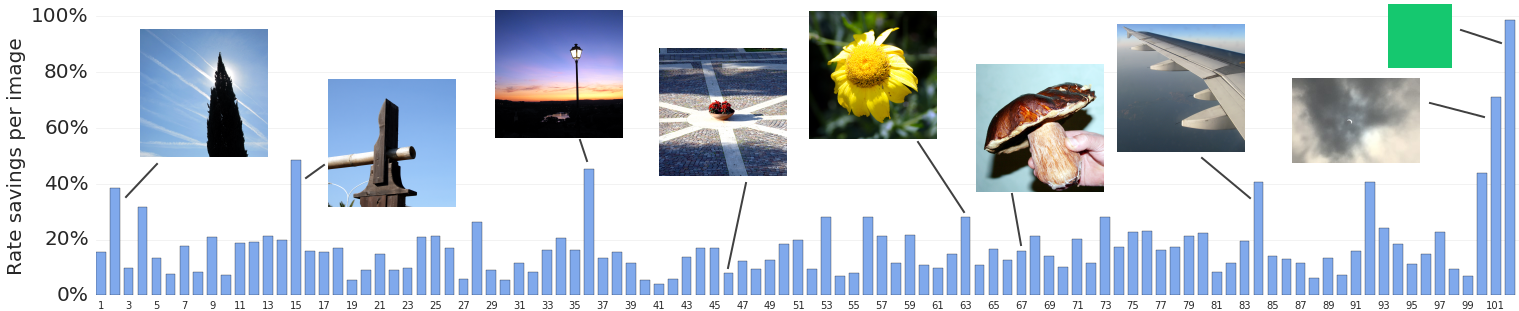}
  \caption{Rate savings for SLIMD over the baseline model (bigger is better) for 100 images from Tecnick plus two additional images showing extreme rate savings (71.2\% for the cloudy eclipse and 98.5\% for the solid green image). Note that SLIMD improves the entropy model without affecting the decoder so the rate savings do \textbf{not} reduce image quality.}
  \label{fig:savings-per-image}
\end{figure*}

The standard approach for ANN-based compression methods is to optimize a fixed entropy model over the training images that is then shared between the encoder and decoder. The form and complexity of the entropy model varies dramatically across different methods including a fully-factorized distribution~\cite{BaLaSi17, ThShCuHu17}, a backward-adaptive, conditional distribution over binarized codes~\cite{RiBo17, LiZuGuZhZh17}, and a learned recurrent network optimized to predict the probability of the next code~\cite{ToOMHwViMi16, JoViMiCoSi17}. The two primary trade-offs between these methods is representational power vs. computational complexity and whether or not the models can be jointly learned with the transform networks, which allows co-adaptation and typically boosts the final rate-distortion performance.

The benefit shared by all of these models is that fixed and backward-adaptive models do not require side information so there is no additional data that must be included in the compressed bitstream. However, it is well-known that carefully crafted side information and forward-adaptation can boost compression rates, and this approach is taken by nearly all modern image compression standards~\cite{ohm2004}. In this paper, we present a forward-adaptive approach for learned image compression based on a \underline{s}patially \underline{l}ocal, \underline{i}mage-dependent \underline{m}ultinomial \underline{d}ictionary (SLIMD). We show that SLIMD significantly boosts rate-distortion performance over a state-of-the-art baseline model by as much as 17.8\% averaged over a standard evaluation image set and up to 98.5\% on images with low visual complexity where the code statistics are poorly modeled by pre-learned entropy models (see Section~\ref{sec:eval} for details).

Our method relies on three design choices to increase compression rates. First, we use local instead of global entropy models, which provides a closer match to the observed code statistics. Previous methods were either not locally adaptive at all or used backward-adaptive models conditioned on local context. Second, SLIMD supports arbitrary distributions (unconstrained multinomials) while requiring relatively little side information. Although the codes produced by our encoder network appear to follow a zero-centered Laplacian distribution on average, the local distributions have considerable variation and often have non-zero mean or multiple modes. Finally, our method can estimate and optionally transmit image-dependent entropy models when the encoder determines that the rate savings outweigh the cost of encoding the custom distribution.


%% file: method.tex
\section{Method}
\label{sec:method}

Our method increases compression rates by improving the entropy models, but the architecture and parameters of the deep networks used for nonlinear transform coding are not affected. This allows SLIMD to work with any model that generates a tensor of quantized codes. For evaluation, we use a model with the same structure as~\cite{BaLaSi17} but with increased model capacity: four convolutional layers instead of three, up to 192 filters per transform layer, and up to 320 code channels. We found that the increased capacity improved rate-distortion performance at higher bit rates, and our baseline results at lower bit rates match those in~\cite{BaLaSi17} almost exactly.

\begin{figure*}[t]
  \centering
  \includegraphics[width=0.495\linewidth]{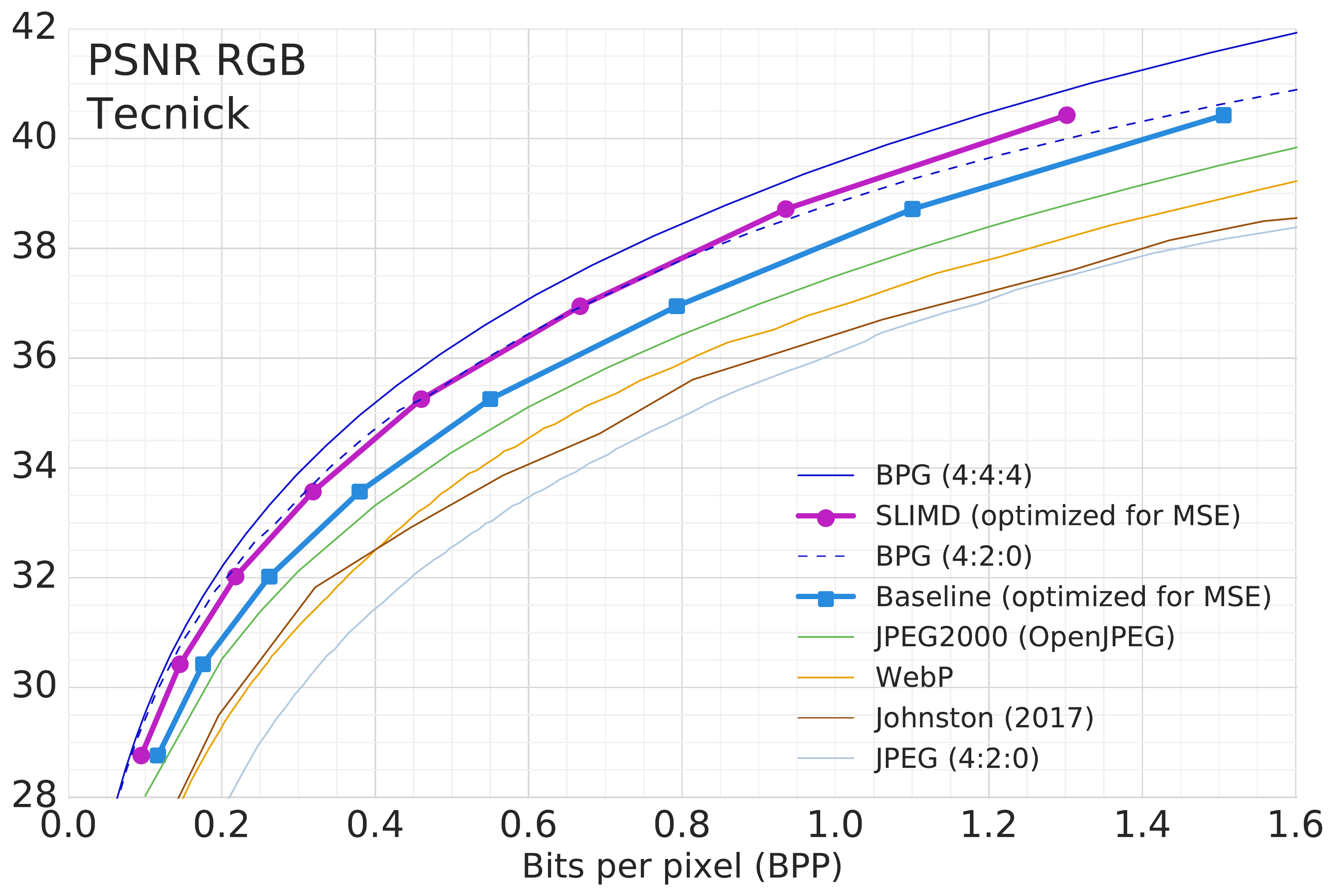}
  \hfill
  \includegraphics[width=0.495\linewidth]{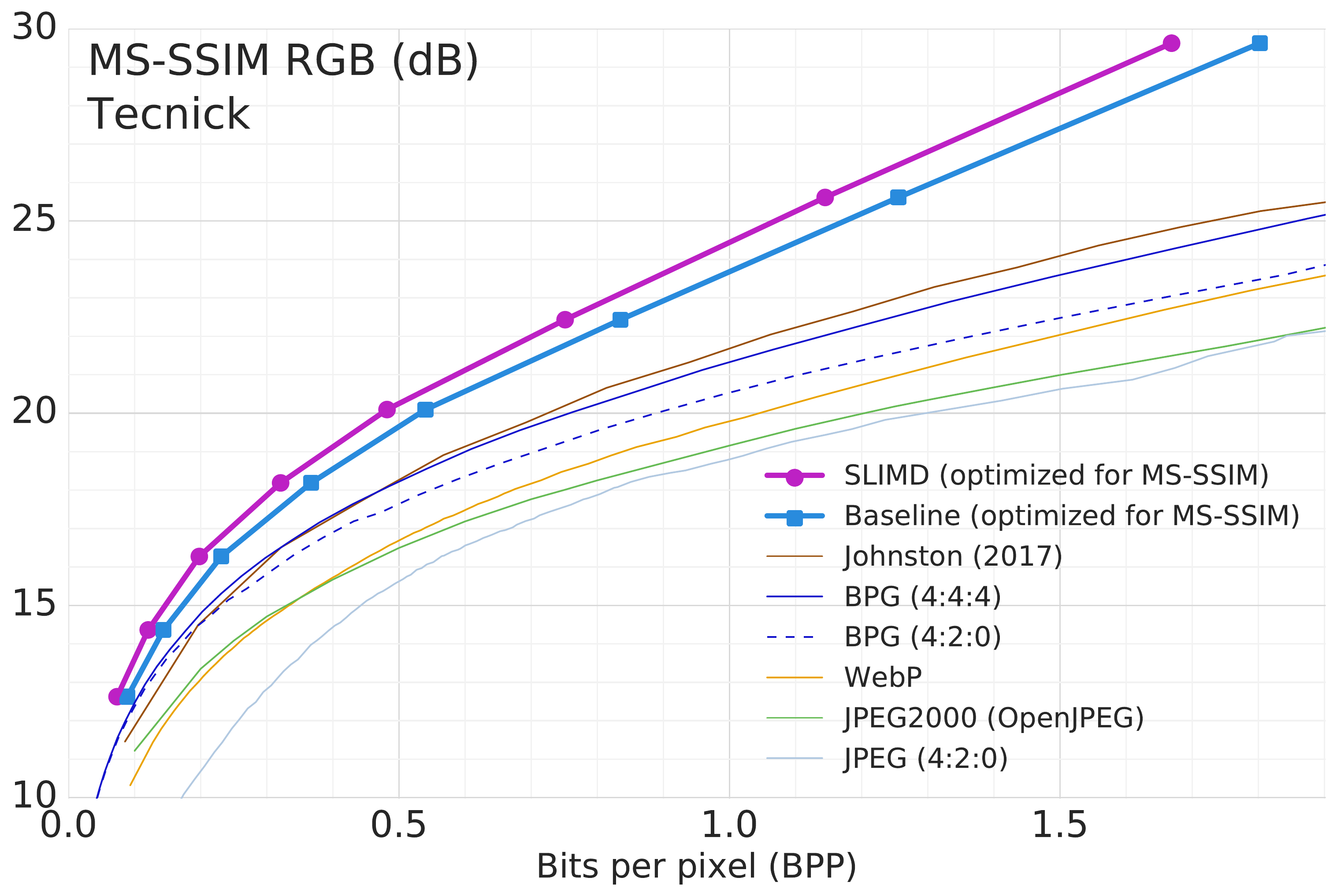}
  \caption{When trained for MSE and evaluated on PSNR (\textit{left}), SLIMD reduces bit rate by an average of 17.88\% over the baseline on the Tecnick image set. SLIMD saves 12.98\% on average for MS-SSIM (\textit{right}).}
  \label{fig:tecnick}
\end{figure*}

\subsection{Local entropy models}
\label{sec:local-models}

Figure~\ref{fig:method} shows a high-level overview of our approach. The first improvement is provided by local entropy models applied by dividing the spatial dimensions of the code tensor into a fixed grid. For instance, for a $1024 \times 1024$ image, our encoder network will create a code tensor with shape [64, 64, 320] (\textit{height, width, channels}), which can then be divided into a $4 \times 4$ grid of $16 \times 16$ tiles. The tile size is a parameter to the encoder and controls the trade-off between better code modeling (smaller tiles) and less side information (larger tiles).

For each spatial tile within each code channel, $T_{y,x,z}$, the encoder would ideally use the true code distribution for entropy coding. However, sending the true distribution as side information is too expensive. Instead, we utilize a clustering algorithm that learns a dictionary of multinomials, $\mathcal{P}$, which allows the encoder to only send the index of the best model for each tile. Given the dictionary, the best model is easily selected by calculating the optimal code length:

\vspace{-2mm}
\begin{equation}
  \text{Index}(T_{y,x,z}) = \argmin_i\: -\sum_{j=1}^n \log_2 p_i(c_j)
  \label{eq:best-distribution}
\end{equation}
 
\noindent
where $p_i \in \mathcal{P}$, $p_i(c_j)$ gives the probability of code $c_j$ using the $i^{\text{\tiny th}}$ entropy model, and $j$ iterates over the $n$ codes in the tile at location $(y,x,z)$ in the 3D code tensor. In practice, we use at most 256 entropy models so the encoder generates one byte of side information for each tile. The model indices are then losslessly compressed before being added to the bitstream (we use DEFLATE~\cite{deflate} for the results in Section~\ref{sec:eval}) so the final cost is typically much less than one byte per tile.

\subsection{Learning a dictionary of entropy models}
\label{sec:learn-entropy-models}

To learn $\mathcal{P}$, the dictionary of multinomials used as local entropy models, we want to minimize the KL-divergence over a training set of code tiles. Specifically, we collect $N$ code tiles from training images, calculate a normalized histogram for each example, $q \in \mathcal{Q} : |\mathcal{Q}| = N$ and $\forall q \sum_i q_i = 1$, and then seek a dictionary that minimizes:

\vspace{-1mm}
\begin{equation}
    \mathcal{L} = \sum_{j=1}^{N} \min_i \KLD{q_j}{p_i}
    \label{eq:cluster-loss}
\end{equation}

\noindent
The result of this minimization is a set of distributions which best code the training data. Finding the global minimum of Eq.~\ref{eq:cluster-loss} is intractable so we optimize using a heuristic method based on the K-means algorithm. First, we run the standard K-means++ algorithm~\cite{kmeans} to get an initial set of distributions. Then we refine the result with a modified version of K-means that uses KL-divergence instead of the $L_2$-norm to map data points to the closest cluster center. This generates a satisfactory set of clusters, except that some may never be used. In order to alleviate this problem, after each full update of K-means, the centroids which have no assignments are updated by replacing them with a random training sample. The result is a dictionary of $N$ multinomials from which the encoder can select the best entry for each code tile.

\begin{figure*}[t]
  \centering
  \includegraphics[width=0.495\linewidth]{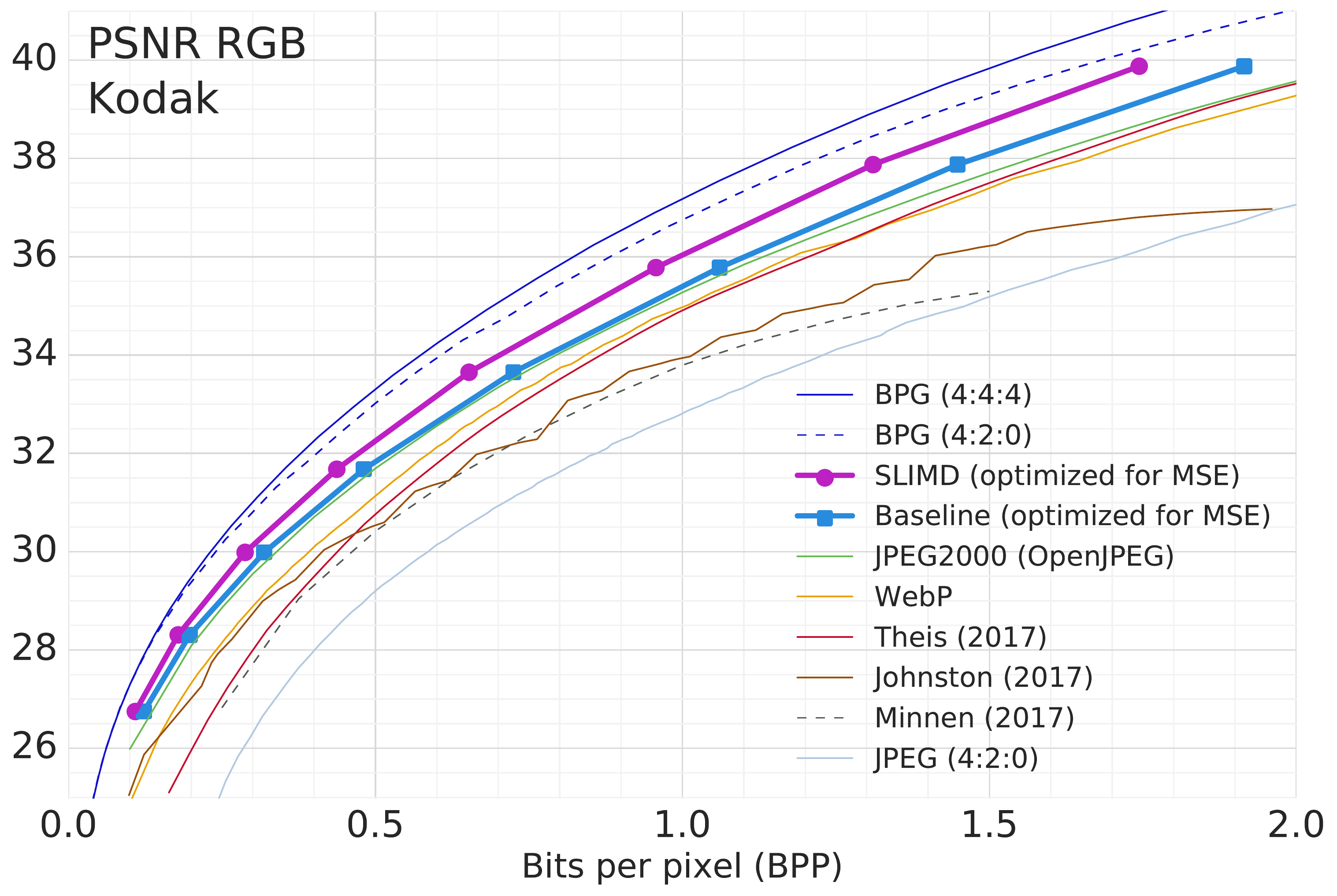}
  \hfill
  \includegraphics[width=0.495\linewidth]{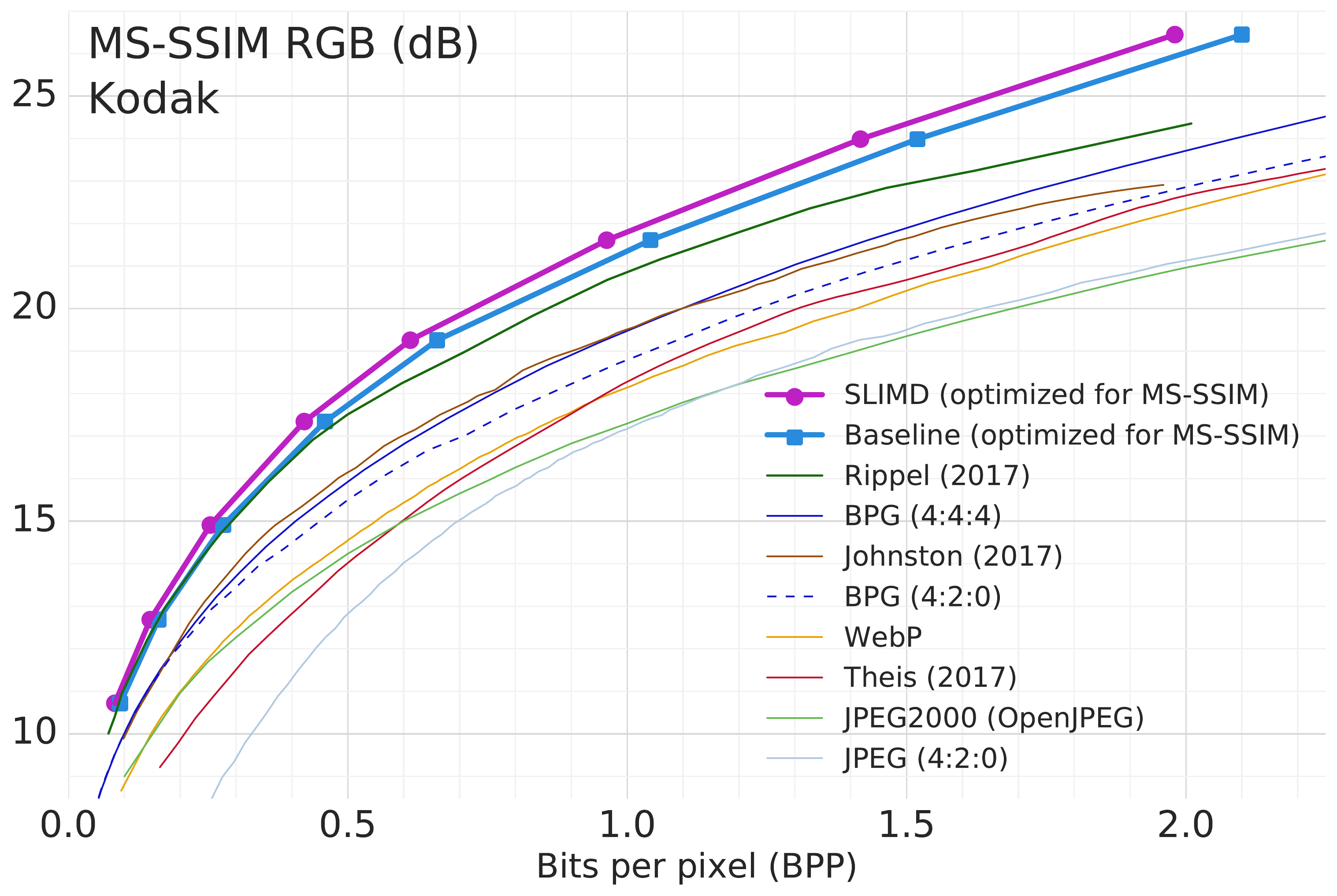}
  \caption{When trained for MSE and evaluated on PSNR (\textit{left}), SLIMD reduces bit rate by an average of 9.29\% over the baseline on the Kodak image set. SLIMD saves 7.67\% on average for MS-SSIM (\textit{right}). SLIMD also outperforms the previous best reported MS-SSIM result~\cite{RiBo17} by 16.9\% on average.}
  \label{fig:kodak}
\end{figure*}

\subsection{Image-dependent entropy models}
\label{sec:image-dependent-models}

The dictionary of local entropy models significantly improves compression rates compared to a single, global model. However, the learned distributions are still optimized to efficiently code images on average (the training set acts as a sample from the distribution of natural images) so further benefits are possible by sending image-dependent models.

Ideally, the encoder would solve an optimization problem that determines the best set of image-dependent distributions while accounting for the cost of transmitting the custom models as additional side information. We believe that such an optimization would be prohibitively slow so our method is constrained to send at most one custom distribution per code channel. To estimate this distribution, the encoder determines which tiles are poorly modeled by comparing the code length under the best dictionary distribution with the code length for the true distribution. If the difference is more than 0.5\%, the tile is included in an average that yields the per-channel custom distribution.

Finally, for each code channel, the encoder calculates the total number of bits saved if the custom distribution is used. If this value is greater than the cost of sending the distribution (estimated as the number of bins between the first and last non-zero element multiplied by 7 bits), the custom distribution is included in the bitstream after being quantized using 8 bit unsigned integers. We use at most 255 pre-learned distributions (0..254) so any tile that benefits from the image-dependent distribution is updated to use index 255.

%% file: eval.tex
\section{Empirical Evaluation}
\label{sec:eval}

The network architecture is the same for all experiments, but we trained two versions of the baseline model: one using mean squared error (MSE) and one using MS-SSIM as the objective function. Results are shown for the model trained with the loss that matches the evaluation metric. We believe that the development of a loss function (or a more general training procedure, \textit{e.g.}, adversarial loss as in~\cite{RiBo17}) that yields state-of-the-art performance across a wide range of perceptual quality metrics is a vital research area, but one that is beyond the scope of this work. Regardless, SLIMD boosts compression rates without modifying the encoder or decoder networks, and so it has the potential to benefit ANN-based transform coders trained using any loss function.

The baseline model has the same structure as in~\cite{BaLaSi17}, which jointly learns a fully-factorized entropy model along with the network parameters. This model is then shared between the encoder and decoder so no side information is required. In our experiments, we compare this baseline to SLIMD, which uses the same deep networks as nonlinear transforms but replaces the fixed, global entropy model with local, image-dependent model as described in Section~\ref{sec:method}.



\subsection{Kodak \& Tecnick evaluation sets}
\label{sec:kodak}

We evaluate our method on two datasets commonly used within the image compression community: the Tecnick image set~\cite{tecnick} and the Kodak PhotoCD dataset~\cite{kodak}. Tecnick consists of 100 images with a resolution of $1200 \times 1200$ pixels, while Kodak consists of 24 images with a resolution of $768 \times 512$ pixels. For both datasets, we evaluate image quality using peak signal-to-noise ratio (PSNR) and the more perceptually relevant multiscale structural similarity index (MS-SSIM)~\cite{wang2003multiscale}.

Figure~\ref{fig:tecnick} summarizes the rate-distortion performance on PSNR (\textit{left}) and MS-SSIM (\textit{right}) calculated over RGB pixel values on the Tecnick dataset. For MS-SSIM, the baseline model already gives state-of-the-art results, and SLIMD further reduces the bit rate by 12.98\% with no loss in image quality. For PSNR, the baseline model outperforms all of the ANN-based methods as well as all of the conventional codecs that we tested (JPEG, JPEG2000, and WebP~\cite{jpeg,jpeg2000,webp}) except for BPG~\cite{bpg}. SLIMD reduces the bit rate by 17.88\% over the baseline making it competitive with BPG (4:2:0). Rate reduction numbers for individual images in the Tecnick set are provided in Figure~\ref{fig:savings-per-image} along with several example images.

Figure~\ref{fig:kodak} provides rate-distortion curves for the smaller Kodak dataset. Here, SLIMD outperforms the previous state-of-the-art result on MS-SSIM~\cite{RiBo17} with an average bit rate reduction of 16.96\%. Finally, SLIMD provides a rate reduction of 7.67\% for the MS-SSIM models and 9.29\% for PSNR when compared to the baseline model.



\subsection{Low-complexity images}
\label{sec:cloudy-eclipse}

The previous section provides average rate reduction results on two standard evaluation image sets. Here, we highlight the savings provided by SLIMD on images with low visual complexity. The most dramatic result is for a one megapixel solid green image: the baseline model requires 0.467 bpp, while SLIMD requires 0.0065 bpp, yielding a savings of 98.57\% (\textit{i.e.}, the baseline encoding is 70x larger). For a more realistic example, the cloudy eclipse image (shown at the right of Figure~\ref{fig:savings-per-image}) requires 0.438 bpp with the baseline model and 0.126 bpp with SLIMD for a rate savings of 71.2\%.


We believe that the relatively poor compression rate of the baseline method is explained by the use of a fixed, global entropy model. The model learns the distribution of codes over the training set, which was collected to approximate the distribution of ``natural images'' found on the web. It is not surprising that the codes for low-complexity images follow a considerably different distribution than the learned prior, which leads to a high cross-entropy and thus a low compression rate. By supporting image-dependent entropy models, SLIMD is able to use a custom distribution that very closely models the codes for these images.

Although these examples were selected to highlight the extreme rate savings possible with SLIMD over the baseline model, we see significant savings even with independent evaluation sets. For example, SLIMD provides a savings of more than 20\% for 27 out of the 100 images in Tecnick with the largest gain reaching 48.6\%.

%% file: conclusion.tex
\section{Conclusion}
\label{sec:conclusion}

We presented an approach for improving the entropy models for ANN-based image compression methods based on a spatially local, image-dependent multinomial dictionary (SLIMD). Our method provides state-of-the-art rate-distortion performance compared to other ANN-based methods as measured by both PSNR and MS-SSIM on two standard evaluation sets. We believe this shows that existing ANN-based compression methods have not effectively utilized ideas like forward-adaptation and that additional research is needed to understand how to learn better entropy models and incorporate other forms of side information into the network optimization procedure.